**MacBehaviour: An R package for behavioural experimentation on large language models**

Xufeng Duan[1]    Shixuan Li[1]    Zhenguang G. Cai[1,2]

[1] Department of Linguistics and Modern Languages, The Chinese University of Hong Kong

[2] Brain and Mind Institute, The Chinese University of Hong Kong

Author notes:

Correspondence should be addressed to either Xufeng Duan, Department of Linguistics and Modern Languages, The Chinese University of Hong Kong, Shatin, Hong Kong, email: xufeng.duan@link.cuhk.edu.hk, or to Zhenguang G. Cai, Department of Linguistics and Modern Languages, The Chinese University of Hong Kong, Shatin, Hong Kong, email: zhenguangcai@cuhk.edu.hk.

**Abstract**

There has been increasing interest in investigating the behaviours of large language models (LLMs) and LLM-powered chatbots by treating an LLM as a participant in a psychological experiment. We therefore developed an R package called "MacBehaviour" that aims to interact with more than 60 language models in one package (e.g., OpenAI's GPT family, the Claude family, Gemini, Llama family, and open-source models) and streamline the experimental process of LLMs behaviour experiments. The package offers a comprehensive set of functions designed for LLM experiments, covering experiment design, stimuli presentation, model behaviour manipulation, logging response and token probability. To demonstrate the utility and effectiveness of "MacBehaviour," we conducted three validation experiments on three LLMs (GPT-3.5, Llama-2 7B, and Vicuna-1.5 13B) to replicate sound-gender association in LLMs. The results consistently showed that they exhibit human-like tendencies to infer gender from novel personal names based on their phonology, as previously demonstrated (Cai et al., 2023).  In summary, "MacBehaviour" is an R package for machine behaviour studies which offers a user-friendly interface and comprehensive features to simplify and standardize the experimental process.

# 1. Introduction

The rapid advancements in large language models (LLMs) over the past few years have heralded a new era in artificial intelligence (AI). These powerful models, exemplified by systems like GPT-4 (OpenAI et al., 2024), Gemini (Gemini Team et al., 2023), and Llama 2 (Touvron et al., 2023), exhibit remarkable capabilities in natural language understanding, generation, and reasoning. LLMs can engage in open-ended dialogue, answer follow-up questions, and assist with complex tasks like coding and analysis (Tamkin et al., 2021). More critically, scaling up language models has led to emergent abilities, ranging from understanding and generating complex texts to exhibiting problem-solving skills (Jiang, 2023; Webb et al., 2023; Wei et al., 2022). Despite these advances, the full scope of LLMs' capacities and their behaviour remain an area of active research and considerable mystery (Wei et al., 2022).

Current benchmarks of LLMs focused on quantifiable tasks like translation and world knowledge (Hendrycks et al., 2021; Zellers et al., 2019), math (Lewkowycz et al., 2022; Zhou et al., 2023) and coding (Peng et al., 2024); while these tests offer a good understanding of model performance, they are limited in tapping into the black box of these machines, especially how these models behave and the extent to which they resemble human cognition. To address this gap, there is a growing call to study machine behaviour (Rahwan et al., 2019) and investigate the capabilities of LLMs through the lens of psychology (Hagendorff, 2023). By subjecting LLMs to classic paradigms from psycholinguistics and cognitive psychology, researchers can probe the extent to which these models exhibit human-like phenomena.

Studying the machine behaviour of LLMs offers two key benefits. First, by better understanding how these models process language, reason, and generate outputs, we can further our understanding of how LLMs carry out various tasks. For instance, Binz and Schulz (2023),

Dasgupta et al. (2023) and others applied classical psychological experiments, like the Linda problem and Wason selection task, to explore LLMs in judgment and decision-making. Theory of mind in LLMs was examined by Sap et al. (2023) and Trott et al. (2023). The personality of LLMs was assessed by researchers like Miotto et al. (2022) and Karra et al. (2023). In terms of behavioural economics, Horton (2023) conducted experiments with GPT-3, and the conceptual analysis of reasoning in LLMs was tackled by Huang and Chang (2023) and Qiao et al. (2023). These studies not only enrich our understanding of language but also contribute to the broader field of economy and cognitive science, providing insights into the LLM's performance in a specific domain (based on task) and its underlying mechanisms (See Hagendorff, 2023 for a review).

Second, comparing language use between humans and machines has significant theoretical implications for linguistics, psychology, and cognitive science. LLMs serve as unprecedented models of language processing, capable of capturing intricate patterns and relationships in textual data (Manning et al., 2020). By subjecting LLMs to classic psycholinguistic paradigms, such as syntactic ambiguity resolution (Huang et al., 2024), structural priming (Michaelov et al., 2023) and pragmatic understanding (Qiu et al., 2023), we can test the extent to which they exhibit human-like linguistic phenomena (see also Cai et al., 2024). This allows researchers to formulate and test hypotheses about the linguistic representations and processes in LLMs and also draw implications about linguistic cognition in humans (Piantadosi, 2023). Divergences between human and machine language processing can also shed light on the unique aspects of human cognition, such as the roles of embodied experience, social interaction, and inductive biases (Dupoux, 2018).

Indeed, the advancement in AI has led to the study of machine behaviour (Binz & Schulz,

2023; Hagendorff, 2023; Huff & Ulakçı, 2024; Rahwan et al., 2019). Machine behaviour advocates studying AI systems such as LLMs through the lens of behaviour, drawing inspiration from psychology's focus on observable responses to structured stimuli. By treating LLMs as black boxes and probing their behaviour through carefully designed experiments, we can gain insights into their capabilities and limitations without relying on access to their internal representations. This approach democratizes the study of LLMs, allowing researchers from diverse backgrounds to contribute to our understanding of these powerful systems.

      However, studying LLM behaviour presents two key challenges. The first is the lack of standard procedures for studying machine behaviour. This inconsistency makes it difficult to compare results or build on previous research, limiting our collective grasp of LLM behaviour. Researchers employ a range of methods to carry out experiments, but they often fail to provide comprehensive details about their approaches. This includes how they create contexts for the experiments and how previous contexts may affect responses. For instance, teams might vary in how they pose experiment instruction (either within the system prompt or directly) and gather responses (either through ongoing conversation or by collecting answers separately for each item). These methodological differences prevent effective comparison and slow progress towards comprehending model capabilities. Another key challenge is the significant programming skills and computational resources demanded by experimentation with LLMs, which many psychologists find challenging. Recent studies frequently use multiple models to evaluate consistency or replicate research with open-source alternatives. The unique API protocol of models adds complexity, requiring researchers to sift through extensive documentation and increasing the time and effort needed for research projects.

In this paper, we present an open-source R package that streamlines the process of designing and running experiments on LLMs. The R package, "MacBehaviour" (short for Machine Behaviour), provides a user-friendly interface for researchers to interact with more than 60 LLMs, including popular models like the GPT family, the Claude family, and open-source alternatives like the Llama family (see Table 1). MacBehaviour abstracts away the technical complexities of interfacing with these models, allowing users to focus on the substantive aspects of their experiments. The package offers a suite of functions for common experimental tasks, such as sending prompts, experiment design, and logging response and token probability. It supports both cloud-based LLMs (access through platforms' APIs, like the GPT family) and self-hosted LLMs via FastChat (Zheng et al., 2023).

**Table 1**. A sample of supported models

| Model | Developer/Platform |
|---|---|
| GPT family (GPT-3.5, GPT-4 et al.) | OpenAI (OpenAI et al., 2024) |
| Claude family (Haiku, Sonnet, Opu et al.) | Anthropic (Anthropic, 2023) |
| Gemini family (Ultra, Pro, and Nano et al.) | Google (Gemini Team et al., 2023) |
| Llama family (Llama-2, Llama-3) | Meta (Touvron et al., 2023) |
| BaiChuan family (7B, 13B et al) | Baichuan Intelligent Technology (Yang et al., 2023) |
| 50+ other self-hosted LLMs (e.g., Vicuna, FastChat-T5) | FastChat (Zheng et al., 2023) |

**Note**. For all supported self-hosted LLMs by FastChat, please see https://github.com/lm-sys/FastChat/blob/main/docs/model_support.md.

In the following sections, we provide an overview of MacBehaviour's key features and walk through example use cases. We then carried out a psycholinguistic experiment on some LLMs using different designs and collected different data from the LLMs to demonstrate the validity of the package.

## 2. Methods

### 2.1 Installation and Setup

The "MacBehaviour" R package works with OpenAI's GPT models, Claude family, Llama family and other models that use the OpenAI-compatible API. This facilitates the conduct of behavioral experiments on LLMs (see Table 2 for complete list of functions in the package). The package is not limited to psycholinguistic experimentation on LLMs but can be used for behavioural investigation of LLMs in general (e.g., decision making, Binz & Schulz, 2023; stimulus norming, Alzahrani, 2024).

**Table 2**. "MacBehaviour" R package main functions

| Function | Description |
| --- | --- |
| **setKey** | Set the API key and URL for an LLM. |
| **loadData** | Load experimental stimuli. |
| **experimentDesign** | Define experiment setup. |
| **preCheck** | Configure model parameters and check the token number of stimuli before execution. |
| **runExperiment** | Execute an experiment and log model responses. |
| **handle_error_code** | Interpret HTTP status codes returned and print explanations for each known status code to the console. |
| **llama_chat** | This internal function sends requests via the hosted Hugging Face API to initiate a conversation with llama-2. |

| **openai_chat** | This internal function sends requests via the OpenAI API to initiate a conversation with GPT family based on multiple prompts. |
|---|---|
| **tokenCheckOne** | This internal function calculates the token count for each prompt in single-trial runs. |
| **tokenCheckRun** | This internal function calculates the token count for prompts across multiple trials within a run. |
| **run_openai** | This internal function manages the execution of the experiment based on the gptConfig settings. It iteratively processes the stimuli for each trial and interacts with OpenAI GPT family or self-hosted models with OpenAI-compatible API models. It is utilized within the 'runExperiment' function. |
| **run_llama** | This internal function manages the execution of the experiment based on the gptConfig settings. It iteratively processes the stimuli for each trial and interacts with the Llama-2 models. It is utilized within the "runExperiment" function. |
| **magicTokenizer** | This function provides the number of tokens for a given text list, acting as a wrapper for an internal tokenizer function. |
| **addMessage** | This internal function is used to append a new message (composed of role and content) to an existing list of messages. This is used internally to manage conversations during data collection. |

Users can install this package through "install.package" function or GitHub repository:

``` r
Install.package("MacBehaviour")
```

``` r
install.packages("devtools")
devtools::install_github("xufengduan/MacBehaviour")
```

Upon the successful installation, users can load this package into the current R session:

```r
library("MacBehaviour")
```

After package loading, users will need to verify the model information (such as API, URL, and model version) before the experiment:

```r
setKey(api_key = "YOUR_API_KEY", api_url = "YOUR_MODEL_URL", model = "YOUR_MODEL_TYPE)
```

1) The "api_key" argument in this function requires user' personal API (Application Programming Interface, which enables users to access to language models) from OpenAI, Hugging Face or other companies. Fill in "NA" if users are using a self-hosted model. API enables authenticated access to language models. Researchers interested in obtaining OpenAI API key should first sign up on the OpenAI platform (https://platform.openai.com/). After registration, navigate to user's account settings where user can generate personal API key. Similarly, for Hugging Face models, an API key specific to Hugging Face is required. This can be obtained by creating an account on the Hugging Face platform (https://huggingface.co/). Once you are logged in, access your account settings, and find the "access token" to generate Hugging Face API key. Please note that as the model inference needs GPUs, users may need to pay inference costs to OpenAI (https://openai.com/pricing) or Hugging Face (https://huggingface.co/blog/inference-pro; for using Llama-2 family models).

2) The "api_url" argument, a character vector, specifies the interface domain of the selected model. For experiments using the GPT family, the URLs are documented in OpenAI's API reference (https://platform.openai.com/docs/api-reference/authentication). For Llama-2 models available through Hugging Face, the model's URL can be found in the respective model's repository, such as " https://api-inference.huggingface.co/models/meta-llama/Llama-2-70b-chat-hf". For self-hosted models, please fill this argument with the user's local URL ("for more information, see https://github.com/lm-sys/FastChat/blob/main/docs/openai_api.md).

Here, users can modify how a language model generates responses by adjusting the "api_url". There are two modes for generating output from an LLM: "text completion" and "chat completion" (for details, please see https://platform.openai.com/docs/guides/text-generation/chat-completions-vs-completions). The "text completion" mode requires only a preamble as input, after which the model autonomously generates the remaining text (for GPT-3.5, the api_url for text completion is "https://api.openai.com/v1/completions"; "http://localhost:8000/v1/chat/completions" for self-hosted models). Conversely, "chat completion" is a mode for constructing a conversation between a human user and the language model assistant. Therefore, this approach requires a clear definition of roles (assistant vs. user) and a specific prompt for the model to follow for a task (e.g., "Please complete the following preamble..." for text completion). To engage GPT-3.5 in chat completion mode, use the URL https://api.openai.com/v1/chat/completions. For the self-hosted model, access "http://localhost:8000/v1/completions".

3) The "model" argument, a character vector, specifies the index of the selected model. For OpenAI models, you can find the list of available model indexes here: (https://platform.openai.com/account/limits). For self-hosted models, users can find the model's name at the model's corresponding repository (for a summary, see https://github.com/lm-sys/FastChat/blob/main/docs/model_support.md).

## 2.2 Experiment design

"MacBehaviour" can implement an experiment in two types of designs. Firstly, a multiple-trials-per-run design resembles typical psychological experiments, where a human participant encounters multiple trials in an experiment. Here, the user can present multiple experimental trials, one by one, to an LLM in a single run (i.e., conversation). Note that earlier input and output will serve as the context for a current trial. The multiple-trials-per-run design can be used if one intends to strictly replicate a human experiment (as psychological experiments involving humans tend to have multiple trials for each participant) or is interested in investigating the effect of previous experimental trials. Secondly, in a one-trial-per-run design, users only present a single trial of prompt and stimulus to an LLM in a conversation and present another trial in a new conversation; this resembles a human experimental design where a participant comes in just to do one of the trials.

We next use an example study to illustrate these designs in this package. Cassidy et al. (1999) showed that speakers of English can infer the gender of novel personal names from phonology. In particular, when asked to complete a sentence fragment (e.g., *After Corlak/Corla went to bed ...*), people tend to use a masculine pronoun for names ending in a closed syllable (e.g., *Corlak*) but a feminine pronoun for those ending in an open syllable (e.g., *Corla*). Cai et al. (2023) replicated the experiment with ChatGPT and Vicuna and obtained a similar phonology-gender association in these LLMs. In the following parts, we show how to use the "MacBehaviour" package, using this experiment as an example. Following Cai et al. (2023), in our demo, we ask an LLM to complete sentence fragments and observe how the model refers to the novel personal name (e.g., using masculine pronouns such as *he/him/his* or feminine ones such as *she/her/hers*).

*2.2.1 Multiple-trials-per-run design*

Before using this package, users should prepare one CSV file/data frame containing the experimental stimuli and other information for experiment design (see Table 3). The CSV file/data frame should exhibit a structured format, defining columns for "Run", "Item", "Condition", and "Prompt", with each row standing for a unique stimulus (see Table 3 for a description of these terms and Table 4 for an example). This organization is pivotal for keeping the integrity of the experimental design, ensuring that each stimulus is correctly identified and presented to an LLM according to user's experiment design.

**Table 3**. The data frame structure

| Column | Description |
| --- | --- |
| **Run** | Index of the conversation with the model, akin to the concept of "list" in a psychological experiment. Items shared with the same Run index will be presented in a single conversation. |
| **Item** | Index of stimuli for data tracking and organization. |
| **Condition** | The experimental condition associated with each stimulus, for researcher's reference. |
| **Prompt** | The actual prompt, together with a stimulus, presented to the model |

**Note.** Each row stands for a unique stimulus in the data frame/sheet.

**Table 4**. An exemplar stimuli file in a multiple-trials-per-run design

| Run | Item | Condition | Prompt |
|---|---|---|---|
| 1 | 1 | Open syllable | Please repeat the fragment and complete it into a full sentence: Although Pelcra was sick … |
| 1 | 2 | Closed syllable | Please repeat the fragment and complete it into a full sentence: Because Steban was very careless … |
| 1 | 3 | Open syllable | Please repeat the fragment and complete it into a full sentence: When Hispa was going to work … |
| 1 | 4 | Closed syllable | Please repeat the fragment and complete it into a full sentence: Before Bonteed went to college … |
| 1 | 5 | … | … |
| 1 | … | … | … |
| 2 | 1 | Closed syllable | Please repeat the fragment and complete it into a full sentence: Although Pelcrad was sick … |
| 2 | 2 | Open syllable | Please repeat the fragment and complete it into a full sentence: Because Steba was very careless … |
| 2 | 3 | Closed syllable | Please repeat the fragment and complete it into a full sentence: When Hispad was going to work … |
| 2 | 4 | Open syllable | Please repeat the fragment and complete it into a full sentence: Before Bontee went to college … |
| 2 | 5 | … | … |
| 2 | … | … | … |

In the multiple-trials-per-run design (see Table 4), multiple trials (e.g., 32 trials in our validation study 1 below) are presented in a single conversation (run). In each Run, the package sends the stimulus based on the index of the row. Users can randomize the item order within Runs in the function "experimentDesign" later. The LLM takes the prompt and the stimulus as input, with input and model output in earlier trials as its context.

To achieve the above conversation, this package sends the stimuli in the following format for OpenAI GPT family:

```r
# OpenAI/Open-source models
# For the first trial:
[list(role = "user", content = "Please repeat the fragment and complete it into a full sentence: Although Pelcra was sick …")]

# For the second trial:
[list(role = "user", content = "Please repeat the fragment and complete it into a full sentence: Although Pelcra was sick …"),
list(role = "assistant", content = "Although Pelcra was sick, she remained determined to finish her project on time. "),
list(role = "user", content = "Please repeat the fragment and complete it into a full sentence: Because Steban was very careless …")]
```

In this context, there are two roles: "user" and "assistant" (identified as "ChatGPT," tasked with providing responses). The conversational context was provided at the beginning of the next trial's prompt. In this example, the context included the first stimulus *Please repeat the fragment and complete it into a full sentence: Although Pelcra was sick ...* and its response *Although Pelcra was sick, she remained determined to finish her project on time.* The prompt then presented the second stimulus *Please repeat the fragment and complete it into a full sentence: Because Steban was very careless ...* after the conversational context. We implemented this function for other cloud-based models and self-hosted models in the same way (see more at

https://Huggingface.co/blog/llama2#how-to-prompt-llama-2; https://github.com/lm-sys/FastChat/blob/main/fastchat/conversation.py).

*2.2.2 One-trial-per-run design*

In the one-trial-per-run design, an LLM will be presented with only one trial of the experiment in a run/conversation. In our demo design here (see Table 5), each conversation with the LLM involves only one stimulus. In this design, each stimulus is given a unique run number, indicating that each one is to be presented in a separate conversation with the LLM. This design eliminates the potential for the previous context to influence the response of the current stimulus, ensuring that each stimulus is evaluated independently.

**Table 5**. Stimuli for one-trial-per-run design

| Run | Item | Condition | Prompt |
| --- | --- | --- | --- |
| 1 | 1 | Open syllable | Please repeat the fragment and complete it into a full sentence: Although Pelcra was sick … |
| 2 | 1 | Closed syllable | Please repeat the fragment and complete it into a full sentence: Although Pelcrad was sick … |
| 3 | 2 | Open syllable | Please repeat the fragment and complete it into a full sentence: Because Steba was very careless … |
| 4 | 2 | Closed syllable | Please repeat the fragment and complete it into a full sentence: Because Steban was very careless … |
| 5 | 3 | Open syllable | Please repeat the fragment and complete it into a full sentence: When Hispa was going to work … |
| 6 | 3 | Closed syllable | Please repeat the fragment and complete it into a full sentence: When Hispad was going to work … |
| 7 | 4 | Open syllable | Please repeat the fragment and complete it into a full sentence: Before Bontee went to college … |

| 8 | 4 | Closed syllable | Please repeat the fragment and complete it into a full sentence: Before Bonteed went to college … |
|---|---|---|---|
| 9 | 5 | … | … |
| 9 | 5 | … | … |

## 2.3 Experimental Pipeline

*2.3.1 Standardizing stimuli*

```r
df = read.csv("/path/to/CSV/demo.csv")
```

The "read.csv" reads the CSV file, converting it into a data frame within R. Users can also import a data frame containing stimuli and experiment information through other functions. To accurately present the stimuli within the R environment, the "loadData" function is utilized, which organizes the data from a data frame for further processing:

```r
ExperimentItem = loadData(runList=df$Run, itemList=df$Item, conditionList=df$Condition, promptList=df$Prompt)
```

The "loadData" function maps vectors or data frame columns to specific keywords. These keywords are then recognized by subsequent functions in our framework. This mapping streamlines the automatic identification and processing of relevant data collection:

1) The "runList", required, a numeric vector, matches the column for "Run" in the CSV file and denotes the conversation/run index. It is utilized in loops for interactions with LLMs. The vector's name (e.g., df$Run) can be arbitrary; what's important is the content specified by users for the runList. This applies to subsequent parameters in this function as well.

2) The "itemList", required, a numeric vector, refers to the column for "Item", indicating the item index of stimuli. This is for the researcher's reference and does not interact with the model's operation. It will be used in loops for interactions with LLMs.

3) The "conditionList", required, a numeric/character vector, represents the column for "Condition", which specifies the experimental condition associated with each stimulus. Similar to "itemList", it is for the researcher's reference and does not interact with the model's operation.

4) The "promptList", required, a character vector, maps to the column for "Prompt", which contains the actual prompts that will be presented to the model during the experiment. Each element under this column is a unique prompt the language model will process and respond to.

This package can also interface with models that support multimodal input, such as GPT-4V (https://platform.openai.com/docs/guides/vision) and llava (Liu et al., 2023). For multimodal models, use the labels <text>, <audio>, and  to indicate text prompts, audio inputs, and image inputs respectively. End these with </text>, </audio>, and </img>. For online models like GPT-4V, include the picture download URL; for self-hosted models like llava, users can also use the picture file path. If the study doesn't involve input other than text, simply input the text stimuli without using the <text> label.

The output of this function, "ExperimentItem", is a data frame generated by "loadData", which includes all the necessary details for each stimulus. The accuracy of "loadData" in mapping the CSV spreadsheet/data frame to the "ExperimentItem" is of pivotal importance, as it ensures that each stimulus is precisely presented according to the experimental design.

### 2.3.2 Specifying experimental design

Next, the "experimentDesign" function allows users to define the structure and sequence of the experimental runs (conversations):

```r
Design = experimentDesign(ExperimentItem, Session = 1, randomItem = F)
```

1) "ExperimentItem", required, a data frame, is the output of function "loadData", which is a structured data frame for storing stimuli and experimental information (e.g., item, condition, and prompt for each stimulus).

2) The "Session", optional, an integer, specifies the number of iterations for all stimuli. The default value is 1. It adds a new column named "Session" to your data frame, where each session includes all original stimuli. If the "Session" is set to 2, the package collects data for one session and then repeats all stimuli for a second session.

3) "randomItem", optional, a logical vector, is available to randomize the order of item presentation within a run (conversation). It should automatically remain "FALSE" for the one-trial-per-run design.

### 2.3.3 Model parameters

The model parameters are configured to guide the behaviour of the model during the experiment in the "preCheck" function:

```r
gptConfig = preCheck (data = Design, checkToken = F, systemPrompt = "You are a participant in a psycholinguistic experiment", max_tokens = 500, temperature = 0.7, n = 1, logprobs = True,
)
```

1) "data", required, a data frame, is the output of experimentDesign function.

2) The "systemPrompt", optional, a character vector, offers a task instruction to the model analogous to the instructions given to participants in a psychological experiment. Should one wish to convey the instructions to the model through the trial prompt, one could leave this parameter blank or state some general instructions (e.g., "*You are a participant in a psycholinguistics experiment, please follow the task instruction carefully*"). By default, it is empty. If not, the package will send the systemPrompt content at the start of each run.

```r
[list(role = "system", content = " You are a participant in a psycholinguistics experiment, please follow the task instruction carefully."),
(role = "user", content = "Please repeat the fragment and complete it into a full sentence: Although Pelcra was sick …"),
…]
```

3) The "max_tokens", optional, a numeric vector, limits the length of the model's response. This may lead to an incomplete response if the tokens of response intended by a model exceed this value. The default is 500.

4) The "checkToken", optional, a logical vector, allows users to conduct a token count in order to determine whether their trial(s) have more tokens than a model allows in a single conversation. The default setting, however, is FALSE. When set to TRUE, the package initiates the upload of your experimental stimuli to the tokenizer server of this package for token counting (note that your stimuli will not be retained on the server; they will be promptly removed after the necessary calculations are completed). Our server uses tokenizer algorithms from OpenAI (https://github.com/openai/tiktoken) and Hugging Face (https://github.com/huggingface/transformers/), supporting over 250 models, including OpenAI family, Llama and BERT, automatically selecting the appropriate tokenizer for each. If an unsupported model is chosen, users are alerted with a warning in their report indicating that results were calculated using GPT-2 as the default tokenizer. This ensures transparency about which tokenizer was used, helping users make informed decisions.

For example, consider a study with a one-trial-per-run design that includes 40 items and 100 sessions, where the item with the highest number of tokens has 137. The "checkToken" function generates tailored reports according to your experiment's design. For instance:

```
# One-trial-per-run design
#     CheckItem              Values
# 1   item numbers           4000
# 2   max_token_numbers      137
```

In the report, the "item numbers" show the number of items you have (number of items × number of sessions). The value of "max_token_numbers" signifies the maximum token length among all experimental items. It should not exceed the input token limit of an LLM.

In the report for multiple-trials-per-run design, the package computes the input for the last trial of a run—incorporating all previous conversation history—based on the maximum token count. This is calculated as (systemPrompt + max_tokens) × number of trials + previous conversation history + tokens from the last item; it then reports this total for each run. Please make sure that the max token per run does not exceed the token limit of your selected LLM. The following is an example report.

```
# Multiple-trials-per-run design
# Run         max_tokens_per_run
# 1           1756
# 2           2016
# …
```

5) The "logprobs", optional, a boolean vector, specifies whether to return the log probabilities of output tokens in the chat completion mode. It appends the log probability for each token in the response under the "rawResponse" column. Additionally, users can define how many top probable tokens to display at each token position by introducing a numeric vector "top_logprobs" (https://platform.openai.com/docs/api-reference/chat/create#chat-create-logprobs), which ranges from 0 to 20, showing their corresponding log probabilities. Please note that "logprobs" must be active for this feature to work. Setting it to 2 returns the two most likely tokens at that position. For instance, if "logprobs" is set to TRUE and "top_logprobs" is set to 2, a generated response might be: "Hello! How can I assist you today?" For the first token "Hello", two alternatives are provided:

{"top_logprobs": [{"token": "Hello", "logprob": -0.31725305}, {"token": "Hi", "logprob": -1.3190403}]}

This configuration also provides the two most probable tokens and their respective log probabilities for each subsequent token position.

In the text completion mode (detailed in section "api_url" part in session 2 .1) in the GPT family, "logprobs" is limited to a numeric vector with a maximum value of 5; hence, users don't need to specify candidates by "top_logprobs" (https://platform.openai.com/docs/api-reference/completions/create#completions-create-logprobs). For self-hosted models, currently, only text completion supports collecting token probabilities by setting logprobs to True. This randomly returns one token and its probability at a time, but users can continue requesting until they receive the desired token.

6) imgDetail, optional, offers three settings for image input: low, high, or auto. This allows users to control the model's image processing and textual interpretation. By default, the model operates in "auto" mode, automatically selecting between low and high settings based on the input image size (see more for https://platform.openai.com/docs/guides/vision/low-or-high-fidelity-image-understanding). If inputs do not include images, please skip this parameter.

7) The "temperature", optional, a numeric vector, controls the creativity in LLM's responses (https://platform.openai.com/docs/api-reference/chat/create#chat-create-temperature).

8) The "n", optional, a numeric vector, determines how many unique and independent responses are produced by the model for a single trial. For example, if n = 20, users will get 20 unique responses for each request. However, in a multiple-trials-per-run design, this parameter is automatically disabled to prevent branching conversations (https://platform.openai.com/docs/api-reference/chat/create#chat-create-n).

In addition to the parameters mentioned above, users can also enter optional ones. For reference, you can consult OpenAI's documentation (https://platform.openai.com/docs/api-reference/chat/create) or that of the selected model.

## 2.5 Data collection

The "runExperiment" function is the execution phase of data collection. It initiates the interaction with an LLM based on the specified design and parameters, and iteratively collects responses to the stimuli.

```r
runExperiment (gptConfig, savePath = "./demo.csv")
```

1) "gptConfig", required, is the configuration list object containing all the details of the experiment setup, including the system prompt, the chosen model, maximum tokens, temperature, the number of responses and other parameters. This object is crafted in the preceding steps "preCheck".

2) "savePath", required, is the file path where the experiment results will be saved. This should be an accessible directory on the user's machine with the appropriate write permissions. A file name in the path with either the ".xlsx" or ".csv" extension indicates that its contents are saved in "xlsx" or "csv" format, respectively. These formats are particularly friendly for users who may wish to perform additional data manipulation or visualization within spreadsheet software or import the data into statistical software packages for further analysis.

When "runExperiment" is active, the package sends a prompt to the selected language model, records the model's output, and then moves on to the next stimulus as per the experiment design.

*2.6 Result structure*

Upon the completion of the experiment, the responses are compiled into a file. The output file has the following columns:

**Table 6**. The data structure of the output file.

| Column | Description |
| --- | --- |
| **Session** | Index of the session (each session includes all items). |
| **Run** | Index of the conversation. |
| **Item** | Index of the item. |
| **Trial** | The turn index of a conversation. |
| **Condition** | Details the condition under which the item was presented |
| **Prompt** | Contains the original stimulus content sent to the model. |
| **Response** | The model's text response to the stimulus. |
| **N** | The response index (if asked the model to generate multiple responses at once). |
| **Message** | The actual prompt sending to an LLM. |
| **rawResponse** | The raw response received from the LLM by this package. |

## 3. Validation experiments

In this section, we report several experiments that we run via MacBehaviour as validation studies. As mentioned in Section 2.2, Cassidy et al. (1999) showed that English speakers can infer the gender of novel personal names on the basis of phonology. Specifically, they examined whehter participants referred to a novel name using a masculine or feminine pronoun (e.g., *he/his/him/herself* versus *she/hers/her/herself*) when completing a sentence fragment containing a novel personal name (e.g., *After Corlak/Corla went to bed …*). They showed that participants were more likely to use a feminine pronoun for a novel name ending with an open syllable (e.g., *Corla*) than for one ending with a closed syllable (e.g., *Corlak*). Cai et al. (2023) replicated this sound-gender association in ChatGPT and Vicuna. Next, we report three experiments where we replicated the sound-gender association effect using the MacBehaviour package. The first two experiments collected model completions of sentence fragments (one using the multiple-trial-per-run design and one using the one-trial-per-run design), and the third experiment collected model probabilities for masculine and feminine pronoun continuation of the sentence fragment in a one-trial-per-run design.

*3.1 Experiment 1: Multiple-trials-per-run design*

We adapted materials from Cassidy et al. (1999; Experiments 2), using novel personal names with closed/open syllable endings. We crafted 16 sets of sentence fragments, each containing a novel personal name. We collected data from GPT-3.5 and also open-source models Llama-2 7B and Vicuna 1.5 13B (both deployed locally). Only one parameter was explicitly set: max token = 80. This setting caps the maximum length of generated responses to prevent conversations from exceeding the model's token limit. All other parameters were left at their default. In this experiment, the system prompt states, "You are a participant in a psycholinguistic experiment." We deliver the experimental instructions in the first user's prompt. In subsequent trials, we provide only fragments to the LLM, with the item order being randomized (see below).

```r
[list(role = "system", content = "You are a participant in a psycholinguistic experiment.")
list(role = "user", content = "I'd like to play a sentence completion game with you. I will provide a fragment and I would like you to repeat the fragment and complete it into a full sentence. Each turn's preamble is unrelated to the others. Don't give me a preamble or say anything else. Is that OK?")
list(role = "assistant", content = "That sounds like an interesting game. Sure, let's give it a try.")
list(role = "user", content = "Before Gronday got married …")
list(role = "assistant", content = "Before Gronday got married, he traveled the world to find himself and discover his true passions.") …]
```

In data analysis, we employed logit mixed effects modelling (see the R script and data in https://github.com/xufengduan/MacBehaviour/tree/main/Materials). We determined if an LLM assigned a feminine or masculine gender to the novel name by automatically extracting pronouns (*she/her/hers/herself* or *he/him/his/himself*) from its responses. The dependent variable is the trial-level gender reference (masculine vs. feminine, with masculine as the baseline), and name phonology (closed syllable = -0.5, open syllable = 0.5) serves as the fixed effect. The random effect structure was constructed to account for variability at the item level using forward model comparison. Initially, we examined a basic model with only random intercepts for items and runs. Then, we assessed a more complex model incorporating the fixed effect's slope within items and runs. Should any model fail to converge during comparisons for a specific random effect, we will exclude that random effect to guarantee convergence. The final model included the best random effect structure supported by our data, utilizing an alpha of 0.2 to decide whether to retain a random (Matuschek et al., 2017).

The results indicated a consistent trend across all models, with names ending with an open syllable more frequently associated with feminine genders (see Fig. 1). Indeed, LME analyses revealed a significant effect of sound-gender association in all LLMs, with more feminine pronouns for a novel name ending with an open syllable (e.g., *Corla*) than for one ending with a closed syllable (e.g., *Corlak*) (GPT-3.5: 0.64 vs. 0.34, $\beta = 2.19$, $SE = 0.56$, $z = 9.91$, $p < .001$; Llama-2_7B: 0.38 vs. 0.11, $\beta = 2.38$, $SE = 0.50$, $z = 4.72$, $p < .001$; Vicuna: 0.42 vs. 0.21, $\beta = 1.34$, $SE = 0.21$, $z = 6.34$, $p < .001$). These results thus replicated the findings in Cai et al. (2023) and demonstrated that our package is a valid interface for the collection of LLM responses for behavioural experimentation.

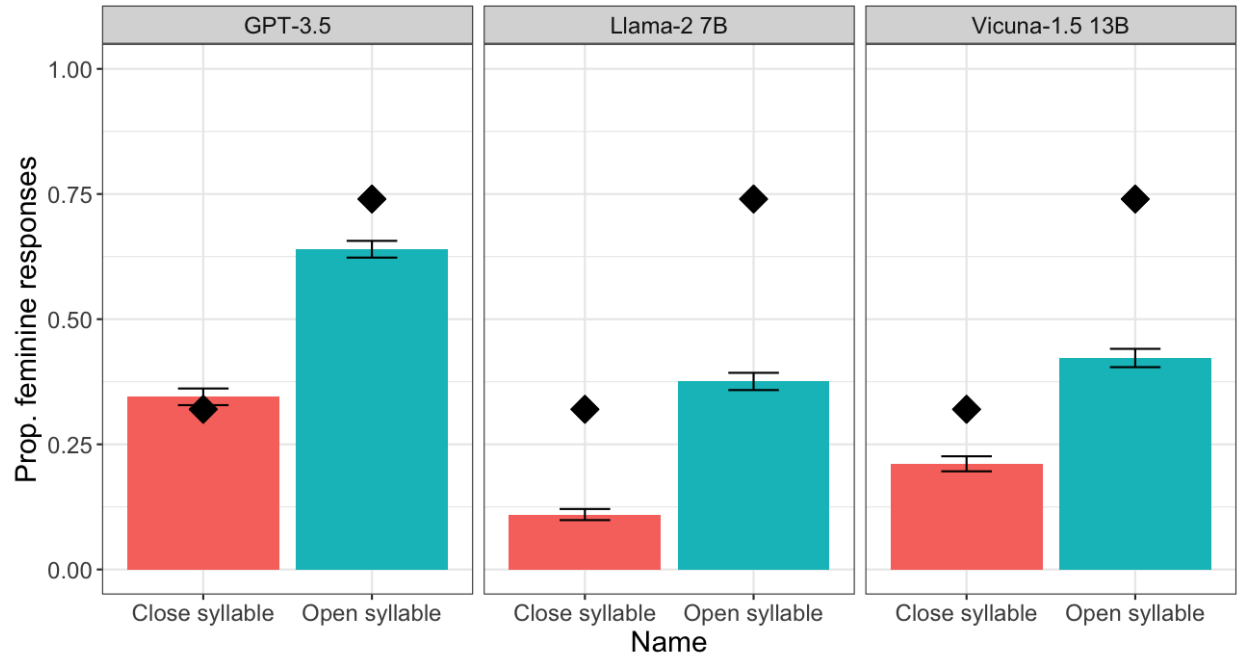

**Figure 1**: Proportion of feminine pronouns for novel names across different LLMs in Experiment 1. Error bars indicate 95% confidence intervals. The diamonds refer to conditional means in Cassidy et al., 1999 (Experiment 2; for reference only as we modified some of the stimuli).

*3.2 Experiment 2: One-trial-per-run design*

This validation experiment was the same as Experiment 1, except that we integrated the experiment instruction and item together, and it used a one-trial-per-run design (see Table 5 and below). All model parameters were left at their default.

```r
[list(role = "system", content = "You are a participant in a psycholinguistic experiment.")
list(role = "user", content = "I'd like to play a sentence completion game with you. I will provide a fragment and I would like you to repeat the fragment and complete it into a full sentence. Although Pelcra was sick …")
list(role = "assistant", content = "Although Pelcra was sick, she still managed to finish her work ahead of schedule.") …]
```

We used the same approach in statistical analyses as in Experiment 1, except that we only considered random effects associated with items; there is no need to consider random effects associated with runs as there was only one trial for each run. Again, as shown in Figure 2, we observed sound-gender association across the LLMs we tested, with more feminine pronouns for a novel name ending with an open syllable than for one with a closed syllable (GPT-3.5: 0.68 vs. 0.08, $\beta = 6.54$, $SE = 1.51$, $z = 4.34$, $p < .001$; Llama-2 7B: 0.63 vs. 0.12, $\beta = 6.54$, $SE = 1.21$, $z = 5.41$, $p < .001$; Vicuna: 0.36 vs. 0.07, $\beta = 3.02$, $SE = 0.50$, $z = 6.08$, $p < .001$).

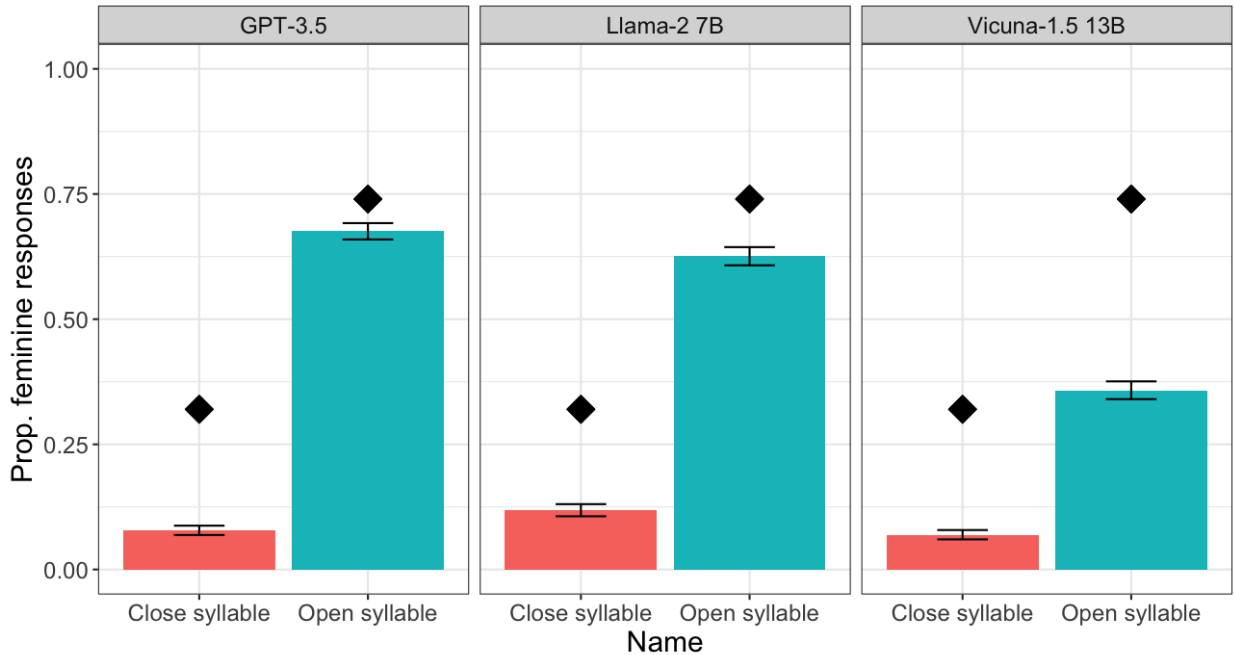

**Figure 2**: Proportion of feminine pronouns for novel names across different LLMs in Experiment 2. Error bars indicate 95% confidence intervals. The diamonds refer to conditional means in Cassidy et al., 1999 (Experiment 2; for reference only as we modified some of the stimuli).

### 3.3 *Experiment 3: Token probabilities*

Previous studies also measure token probabilities using LLMs to represent the likelihood of a word in specific contexts (Goldstein et al., 2022; Huang et al., 2024). In the final validation study, we utilized the "MacBehaviour" package to collect token probabilities in a one-trial-per-run design, employing the same experimental materials as in our previous two studies. We collected data using the text completion mode, where LLMs automatically complete any given input. Therefore, we didn't provide specific instructions for fragment completion or include a system prompt in this mode. We asked LLMs to complete a given sentence fragment and subsequently recorded the probabilities for male (*he/his*) and female (*she/her*) pronouns from the first token produced. We did not consider pronouns such as *him* and *himself/herself* because these are not grammatical continuations for our fragments (e.g., *Although Pelcra was sick…*). For GPT-3.5, we focused on the top five words with the highest probability, adhering to OpenAI's limitation of considering only the first five tokens by setting logprobs = 5 in the "preCheck" function. For the other two self-hosted models, Llama-2 7B and Vicuna-1.5 13B, we collected all pronoun probabilities at the first token position following the fragments.

Our analysis examined the proportions of female and male pronouns, calculated as the percentage of each gender relative to their total probabilities. For example, with females at 30% and males at 40%, the proportion for females is 30% / (30% + 40%) ≈ 42.9%. We examined the difference in proportions between genders for each item using a linear mixed-effects model that incorporated fixed effects for syllable structure (closed syllable = -0.5, open syllable = 0.5) and random effects by item.

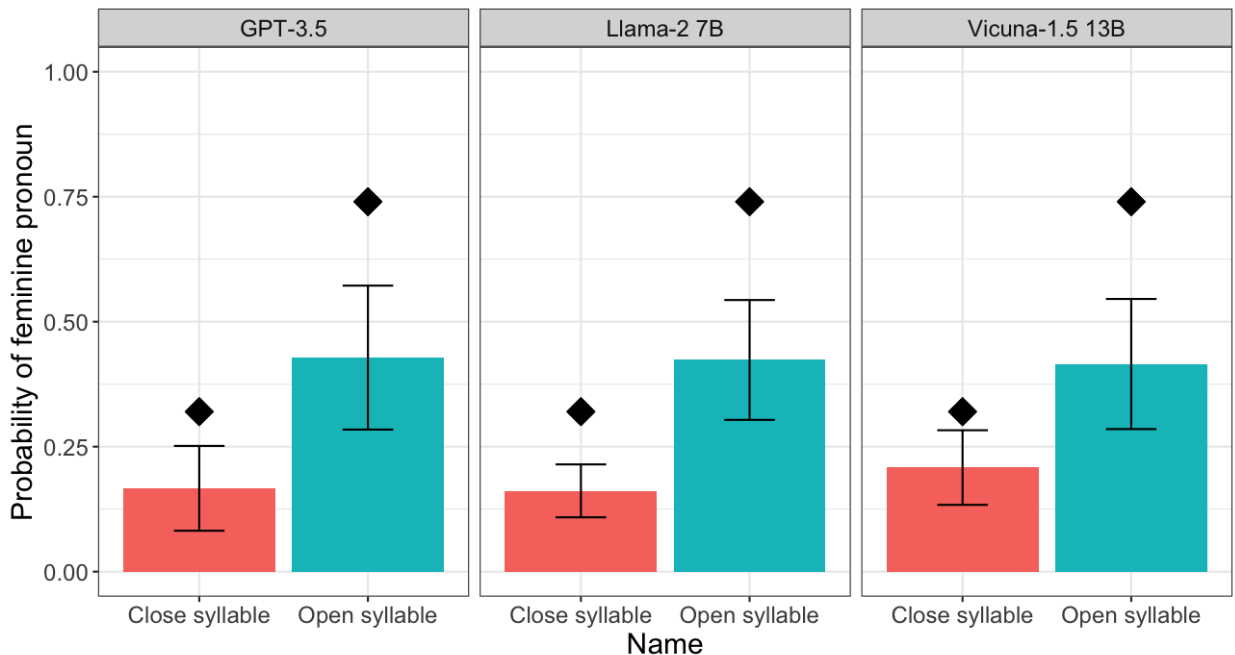

**Figure 3**: Probability results of all models. Error bars indicate 95% confidence intervals. Note that some materials were modified; thus, the human averages are solely for comparison purposes.

The results indicated a significant effect of syllable structure across all models (see Fig. 3). For GPT-3.5, the average proportion difference (female minus male) was higher in open syllables (-0.14) compared to closed syllables (-0.67), resulting in a significant syllable effect ($β = 0.52$, $SE = 0.16$, $t = 3.34$, $p = .002$). The Llama-2 7B model showed a less pronounced, yet significant, difference with -0.15 for open syllables and -0.68 for closed ($β = 0.52$, $SE = 0.12$, $t = 4.35$, $p < .001$). The Vicuna model exhibited a similar trend, with open syllables and closed syllables at -0.58 at -0.17 ($β = 0.41$, $SE = 0.12$, $t = 3.58$, $p = .003$).

To explore the consistency of gender attribution across different experimental conditions, we conducted a correlation analysis between the one-trial-per-run experiment and probability experiments. This analysis focused on comparing the effect of syllable structure on the proportion of female pronoun usage from each model in both experimental paradigms. Specifically, we calculated the difference in female pronoun percentage between open and closed syllables for each item within each model, providing a measure of the phonological effect on gender inference by the language models. The results indicated significant correlations across all models (ChatGPT: $r = 0.59$, $t = 2.74$, $p = 0.016$; Llama-2 7B: $r = 0.43$, $t = 1.77$, $p = 0.099$; Vicuna: $r = 0.62$, $t = 2.98$, $p = 0.010$), suggesting consistency in how models attribute gender based on phonology across different experimental setups. While these correlations indicate a general agreement in model behaviour, the varied strengths of these correlations also reflect underlying differences in how each model processes and responds to novel names in gender attribution across different paradigms.

*3.4 Summary*

Across three distinct validation experiments, we employed identical experimental items (prompts were modified for each experiment) but varied the data collection paradigms to probe the models' ability to sound gender association. This approach highlights the package's versatility and sheds light on how different experimental setups can influence the outcomes even with consistent stimuli.

In the first experiment, we utilized a multiple-trials-per-run design, closely mirroring that in human psychological experiments. This method allowed for the collection of data across several interactions within the same conversation, potentially reflecting cumulative or contextual effects similar to those observed in human subjects. However, this method might also mix effects of individual trials with leftovers from past inputs, adding complexity to the interpretation of single-trial responses. The second experiment adopted a one-trial-per-run design to address the potential confounds noted in the first study. By isolating each trial, this method minimized the influence of the preceding context. The third experiment shifted its focus from generating responses to analyzing the likelihood of pronouns as the next token in completing a sentence fragment. This experiment was unique because it didn't examine the entire sentence generation process but rather focused on the probability of the first token after a prompt. While this approach provides accurate measurements of initial response tendencies, it may not fully capture the models' capabilities since pronouns don't always have to be at the beginning of a sentence and overlooks any efforts in prompt engineering. Also, the results slightly varied among the three experiments, highlighting the importance of considering experimental designs and tasks.

## 4. Discussion

The "MacBehaviour" R package offers an integrated solution for conducting behavioural experiments using LLMs as "participants". It streamlines workflows through easy interaction with various LLMs, including the OpenAI GPT family, Llama-2, self-hosted models and more. While we used a psycholinguistic experiment as a demo here, "MacBehaviour" has a wide array of other potential applications. It could be utilized in the fields of cognitive psychology, behaviour economics, and artificial intelligence. Future updates could enhance more complex experimental paradigms, such as communication between different LLM agents. In conclusion, the "MacBehaviour" R package serves as a methodological toolkit available for research on LLM machine behaviour.

**Open practice statement**

The R package developed as part of this research has been made publicly available on GitHub. The source code can be accessed at https://github.com/xufengduan/MacBehaviour. This package is released under an open-source license, allowing for free use, distribution, and modification. The data and analysis code can be found at https://github.com/xufengduan/MacBehaviour/tree/main/Materials.